\documentclass[letterpaper, 10 pt, conference]{ieeeconf}
\IEEEoverridecommandlockouts                             

\usepackage{pifont}
\usepackage{multirow}
\usepackage{graphicx}
\usepackage{multicol}
\usepackage{booktabs}
\usepackage{amsmath,amssymb}
\usepackage[utf8]{inputenc}
\usepackage[english]{babel}
\usepackage{multirow}
\usepackage[font=small]{caption}
\usepackage{float}
\usepackage[hidelinks]{hyperref}

\DeclareMathOperator*{\argmin}{argmin}
\usepackage{lettrine}
\usepackage[]{algorithm2e}
\usepackage{algpseudocode}
\usepackage{cite}
\usepackage{lipsum}
\usepackage{color}
\usepackage[dvipsnames]{xcolor}
\usepackage{esdiff}
\usepackage{epstopdf}
\usepackage[normalem]{ulem}
\usepackage{soul}

\usepackage{subcaption}
\usepackage{xcolor}
\usepackage{colortbl}
\usepackage{balance}
\newtheorem{defn}{Definition}[section]

\newtheorem{rem}{Remark}[section]

\definecolor{pink}{rgb}{1, 0, 1}
\definecolor{orange}{rgb}{1, 0.7529, 0}
\definecolor{darkgreen}{rgb}{0, 0.8, 0}
\begin{document}

\title{Multi-CAP: A Multi-Robot Connectivity-Aware Hierarchical Coverage Path Planning Algorithm for Unknown Environments}

\author{Zongyuan Shen, Burhanuddin Shirose, Prasanna Sriganesh, Bhaskar Vundurthy, Howie Choset and \\ Matthew Travers
\thanks {All authors are from the Robotics Institute, Carnegie Mellon University, Pittsburgh, USA. {\{zongyuas, bshirose, pkettava, pvundurt, choset, mtravers\}@andrew.cmu.edu}}
\thanks{Supplementary video - \scriptsize\texttt{\url{https://youtu.be/cp83SsG9wjY}}}
}
        
\maketitle

\begin{abstract}

Efficient coordination of multiple robots for coverage of large, unknown environments is a significant challenge that involves minimizing the total coverage path length while reducing inter-robot conflicts. In this paper, we introduce a Multi-robot Connectivity-Aware Planner (Multi-CAP), a hierarchical coverage path planning algorithm that facilitates multi-robot coordination through a novel connectivity-aware approach. The algorithm constructs and dynamically maintains an adjacency graph that represents the environment as a set of connected subareas. Critically, we make the assumption that the environment, while unknown, is bounded. This allows for incremental refinement of the adjacency graph online to ensure its structure represents the physical layout of the space, both in observed and unobserved areas of the map as robots explore the environment. We frame the task of assigning subareas to robots as a Vehicle Routing Problem (VRP), a well-studied problem for finding optimal routes for a fleet of vehicles. This is used to compute disjoint tours that minimize redundant travel, assigning each robot a unique, non-conflicting set of subareas. Each robot then executes its assigned tour, independently adapting its coverage strategy within each subarea to minimize path length based on real-time sensor observations of the subarea. We demonstrate through simulations and multi-robot hardware experiments that Multi-CAP significantly outperforms state-of-the-art methods in key metrics, including coverage time, total path length, and path overlap ratio. Ablation studies further validate the critical role of our connectivity-aware graph and the global tour planner in achieving these performance gains.

\end{abstract}

\section{Introduction}
Coverage Path Planning (CPP) is a capability for autonomous robots with applications in structural inspection, floor cleaning, surveillance, seabed mapping, and robotic weeding~\cite{Veeraraghavan2024,ramesh2024anytime,palacios2019equitable,shen2022ct,shen2025cap}. 
This paper focuses on the challenging case of CPP in unknown environments, which necessitates online methods 
to adapt the coverage paths based on collected environmental information. Multi-robot CPP has gained traction for its ability to accelerate coverage in large-scale environments and improve robustness compared to a single robot~\cite{wang2024apf,tang2025}. 
However, coordinating multiple robots in real time while avoiding inter-robot conflicts and ensuring complete coverage is challenging. This paper introduces a new algorithm to coordinate multiple robots to achieve complete coverage while minimizing inter-robot conflicts and the total coverage path length. The planner incrementally partitions the environment map into multiple subareas, and builds an adjacency graph based on the connectivity between these subareas. At the higher level, the planner assigns each robot a unique subset of subareas to perform coverage. Within each subarea, each robot leverages real-time sensor observation to optimize its coverage path independently. Our experiments show that this strategy helps reduce redundant travel, inter-robot conflicts, and total path length.

\begin{figure}[t]
    \centering
    \includegraphics[width=0.42\textwidth]{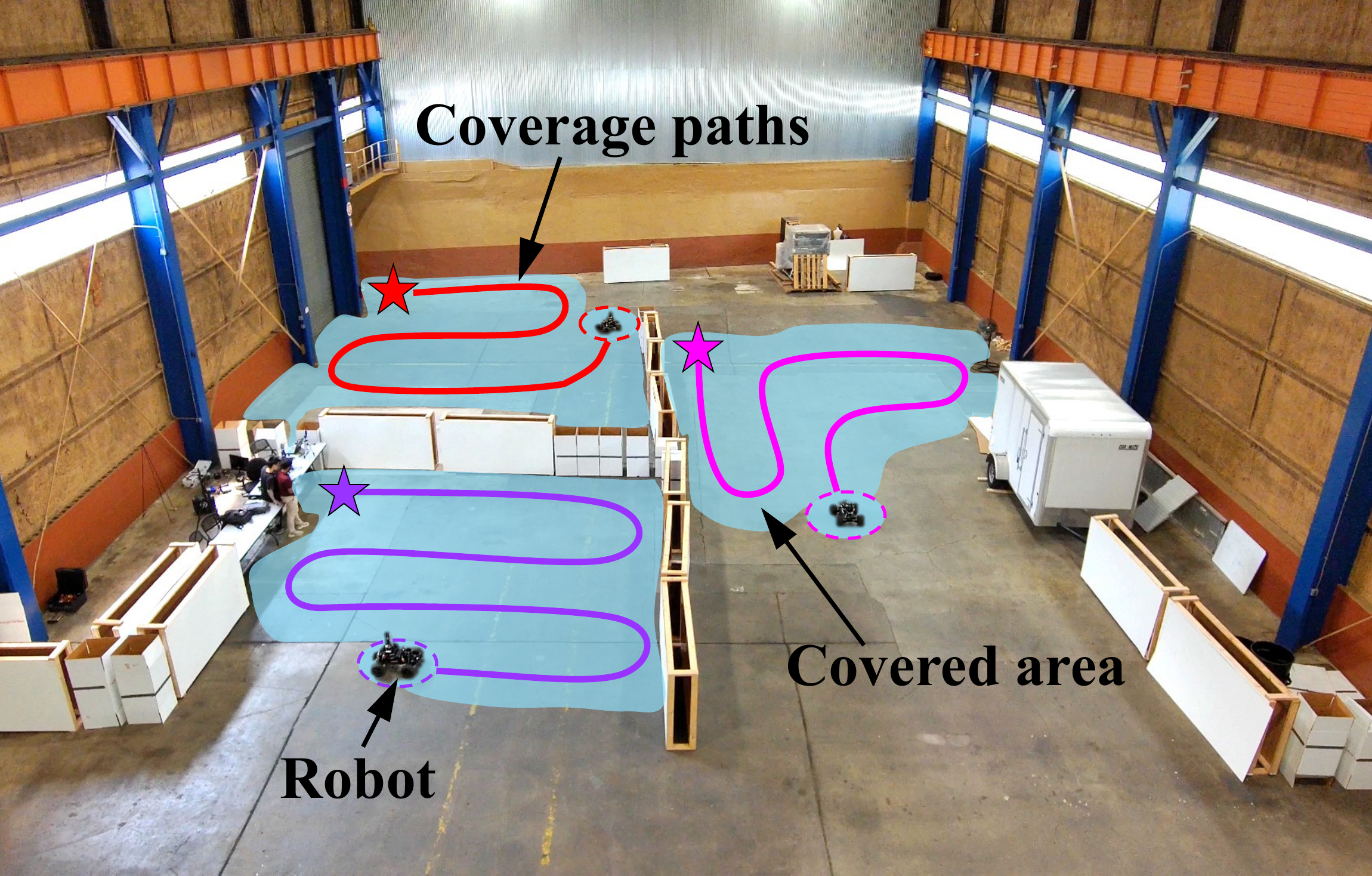}
    \caption{Coverage of a warehouse by using three wheeled robots.}
  \label{fig:coverageexample}
  \vspace{-2em}
\end{figure}

Existing online multi-robot CPP methods face limitations when addressing inter-robot conflicts in cluttered, unknown environments. Some approaches operate by modeling other robots as dynamic obstacles~\cite{luo2017,hassan2020dec,zhang2024herd} or use repulsive potential fields to promote spatial separation between robots~\cite{wang2024apf}. While these approaches minimize conflicts by dispersing the robots, these techniques do not scale well with the number of robots or the complexity of the environment. A different strategy involves simply decomposing the environment into equal-sized subareas and assigning each robot a distinct one~\cite{song2020care,shen2025c,wang2025mac}.
However, these methods decompose the environment without considering obstacle layout. As a result, internal obstacles may split a subarea into disconnected components, forcing the assigned robot to take additional detours during coverage within that subarea. Some methods~\cite{rekleitis2008efficient} incorporate connectivity into the decomposition, but rely on greedy, short-horizon task allocation, leading to workload imbalance and redundant traversal.

To address these challenges of scalability, connectivity, and workload balance, we present the \textbf{Multi}-Robot \textbf{C}onnectivity-\textbf{A}ware Hierarchical Coverage Path \textbf{P}lanning algorithm (\textbf{Multi-CAP}) for unknown environments. The algorithm initializes an adjacency graph by partitioning the environment into a coarse uniform grid. Each grid cell corresponds to a subarea and is represented as a node, while edges encode adjacency relationships between neighboring subareas. This initial subarea may be recursively divided into new subareas as newly discovered obstacles fragment it into disconnected components. As the algorithm operates online, the graph is incrementally refined using sensor feedback to 1) ensure each subarea remains a connected component of non-obstacle cells within the region it originated from, and 2) update adjacency relations to reflect newly discovered obstacles. When obstacles are encountered, the graph is updated by handling blocked edges (removing them if adjacency is lost or recomputing them if an alternate connection remains), excluding obstacle cells from subareas, and splitting disconnected subareas into multiple new ones with corresponding updated edges. This process ensures the graph represents the physical layout of the environment.

We implement a hierarchical strategy using the refined graph. At the higher level, we formulate an Open Multi-Depot Vehicle Routing Problem to compute subarea traversal tours for the robots. These tours are disjoint, assigning each robot a unique set of subareas. This promotes spatial distribution to reduce inter-robot conflicts while simultaneously minimizing overall path length. At the local level, each robot independently adapts its coverage strategy based on the observation status of its assigned subarea. A back-and-forth pattern is applied in partially observed subareas, while a shortest coverage path is computed to cover fully observed ones and ends at the entry cell to the next subarea. This hierarchical strategy integrates global coordination with adaptive local execution, effectively reducing redundant travel and inter-robot conflicts. Our experiments show reduced path length and coverage times compared to baseline algorithms.

\section{Related Work}\label{sec:review}

The multi-robot CPP methods are broadly classified into offline or online. Offline methods assume that a complete map of the environment is available \textit{a priori} and leverage this knowledge to generate globally optimized coverage plans. In contrast, online methods adapt the coverage path in real time based on the collected environmental information. Several offline approaches have shown strong performance in known environments. Palacios et al.~\cite{palacios2019equitable} generate equitable partitions to balance workload across robots. Lu et al.~\cite{lu2023} minimize turning cost by decomposing the map into bricks and optimizing over a bipartite graph. Tang et al.~\cite{tang2023mixed} use a Mixed Integer Programming formulation to optimize coverage time. However, these methods rely on complete prior maps and are unsuitable for unknown environments.

In the online setting, Luo et al.~\cite{luo2017} construct a potential field by assigning negative potentials to obstacles and other robots, and positive potentials to uncovered cells, which then guides the decision-making of individual robots. Hassan et al.~\cite{hassan2020dec} incorporate a collision-avoidance term into the multi-objective cost function for decision-making. Zhang et al.~\cite{zhang2024herd} prioritize coverage near each robot’s starting location, gradually expanding outward. Wang et al.~\cite{wang2024apf} use repulsive forces to promote robot separation during coverage. These methods model other robots as dynamic obstacles~\cite{luo2017,hassan2020dec,zhang2024herd} or use repulsive potential fields~\cite{wang2024apf} to spread out the robots. Since they allow all robots to operate over the entire space, they can lead to inter-robot conflicts and overlapping paths in cluttered environments.

To enhance coverage efficiency and reduce inter-robot conflicts, some methods decompose the environment into subareas and assign each robot to a distinct subarea for coverage. Works in~\cite{song2020care,shen2025c} partition the environment into fixed square subareas, which remain static throughout coverage and are covered using a back-and-forth pattern. Wang et al.~\cite{wang2025mac} dynamically generate the smallest rectangular region enclosing all uncovered cells and then partition it into equal-sized subareas for assignment. However, these methods rely on simple geometric decompositions that fail to capture the environment’s connectivity. As a result, a single subarea may be broken into disconnected parts by internal obstacles, leading to long detours during local coverage. Rekleitis et al.~\cite{rekleitis2008efficient} incorporate connectivity explicitly into the decomposition. The environment is first partitioned into stripes. Robots explore stripe boundaries and detect critical points using the cycle algorithm~\cite{acar2002sensor}, while simultaneously constructing a adjacency graph that incrementally decomposes the environment into subareas. After exploration, these subareas are allocated through greedy distributed auctions based on a heuristic travel cost, and each robot covers its assigned region using a systematic back-and-forth pattern. In contrast, our algorithm defines subareas while explicitly considering the number of robots for balanced workload, allocates the subareas through a global tour planning framework that coordinates both short- and long-term progress, and covers fully observed subareas using the shortest coverage path with specified exit points to avoid retracing already covered areas during transitions. Moreover, our algorithm performs coverage while updating environmental map simultaneously, ensuring continuous progress without distinct phases.
\section{Problem Description}\label{sec:problem_description}

\begin{figure*}[t]
        \centering
    \subfloat[Initialized coverage graph.]{
        \includegraphics[width=0.48\columnwidth]{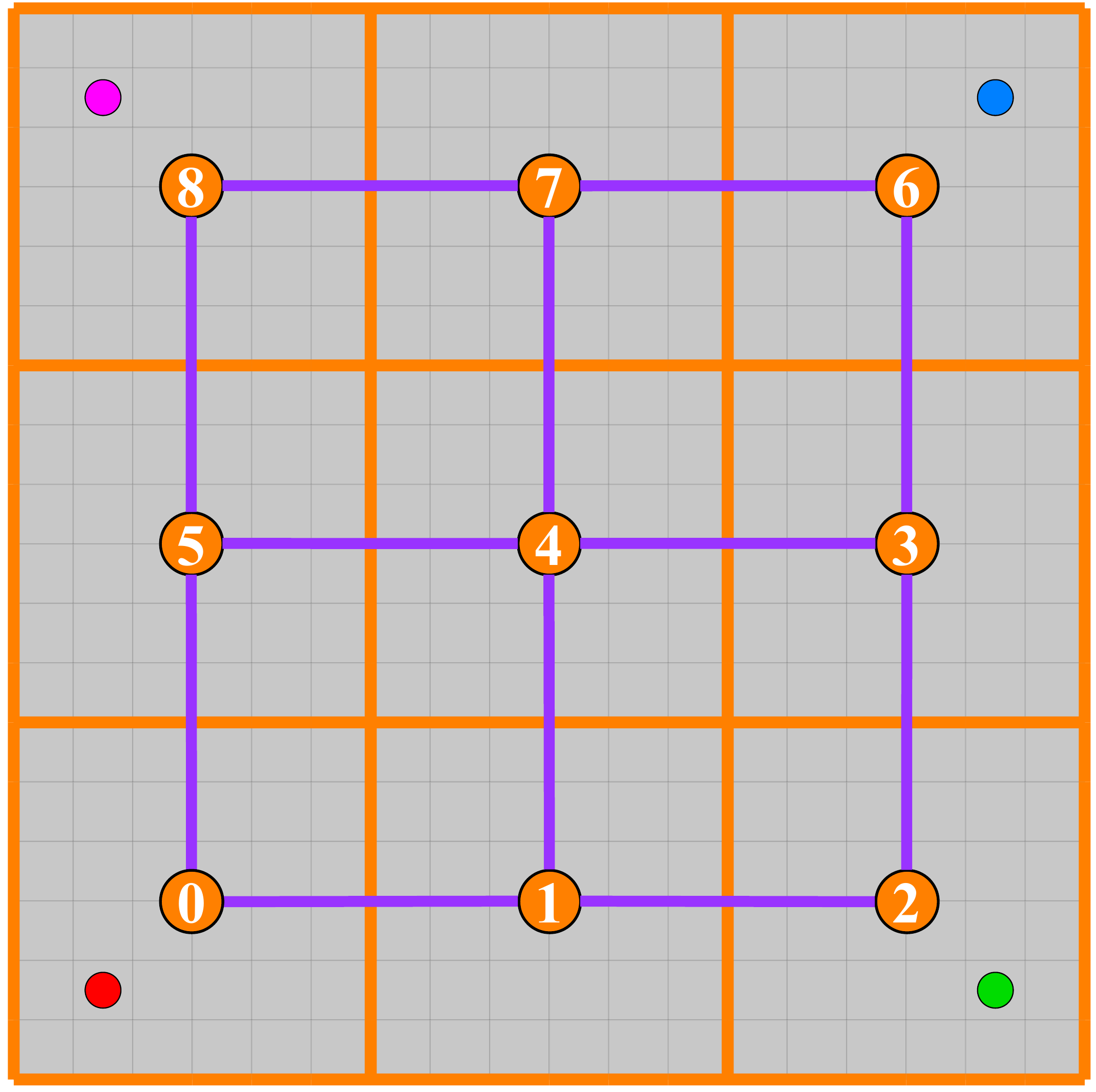}\label{fig:example_part1}}\quad \hspace{-10pt}
        \centering
    \subfloat[Updated coverage graph.]{
        \includegraphics[width=0.48\columnwidth]{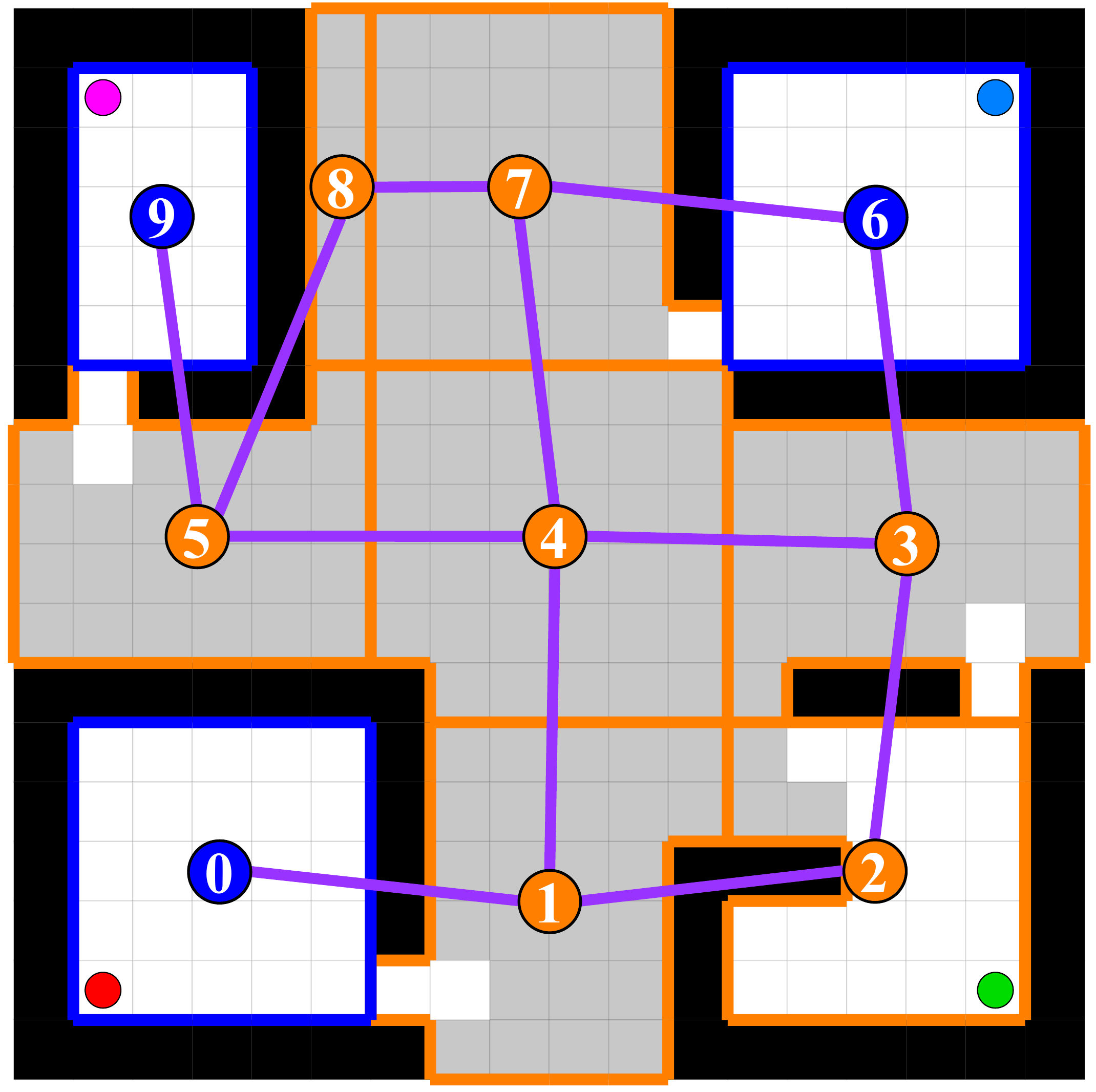}\label{fig:example_part2}}\quad \hspace{-10pt}
        \centering
    \subfloat[Computed global tours]{
        \includegraphics[width=0.48\columnwidth]{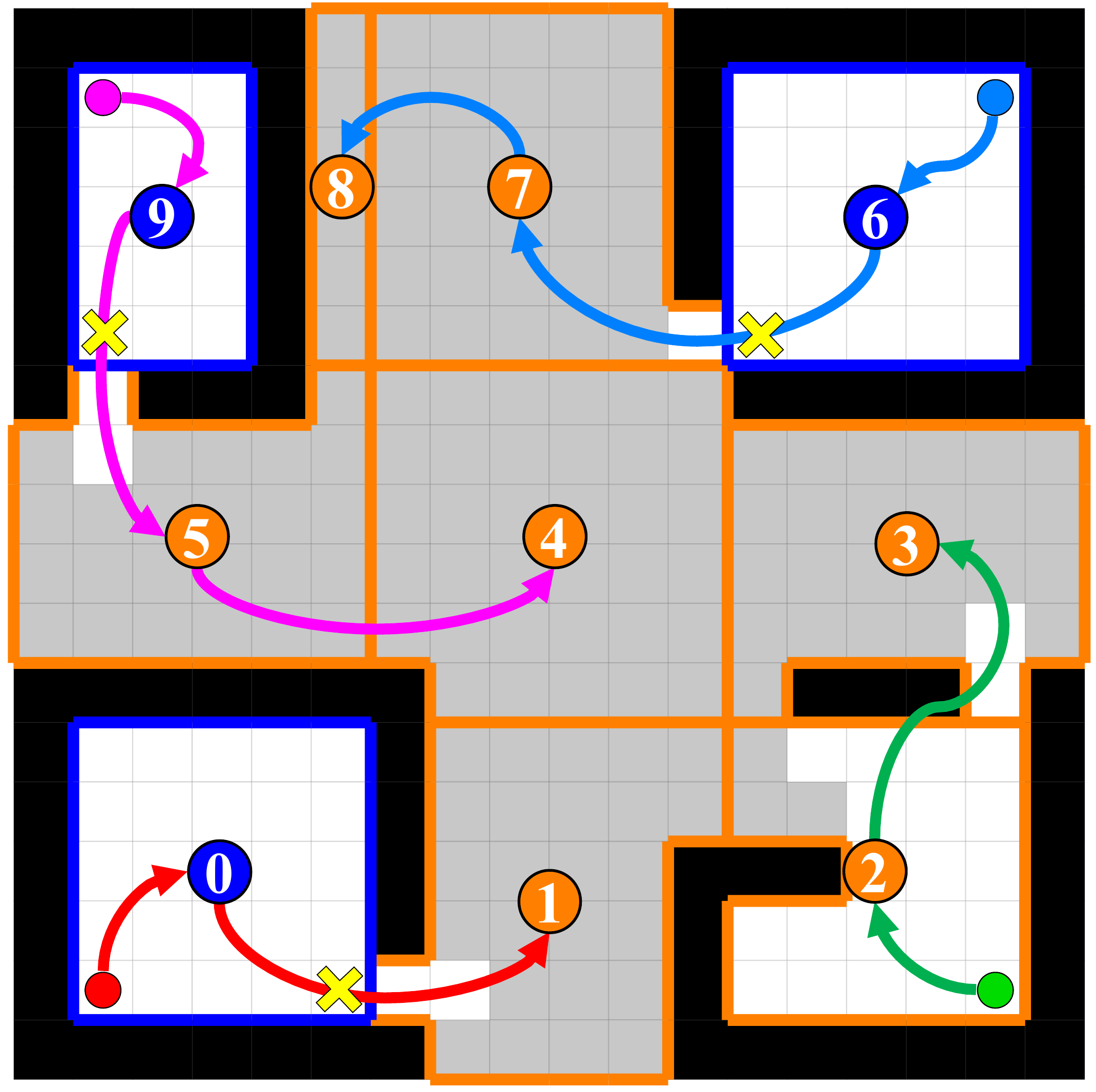}\label{fig:example_part3}}\quad \hspace{-10pt}
        \centering
    \subfloat[Adaptive local coverage paths.]{
        \includegraphics[width=0.48\columnwidth]{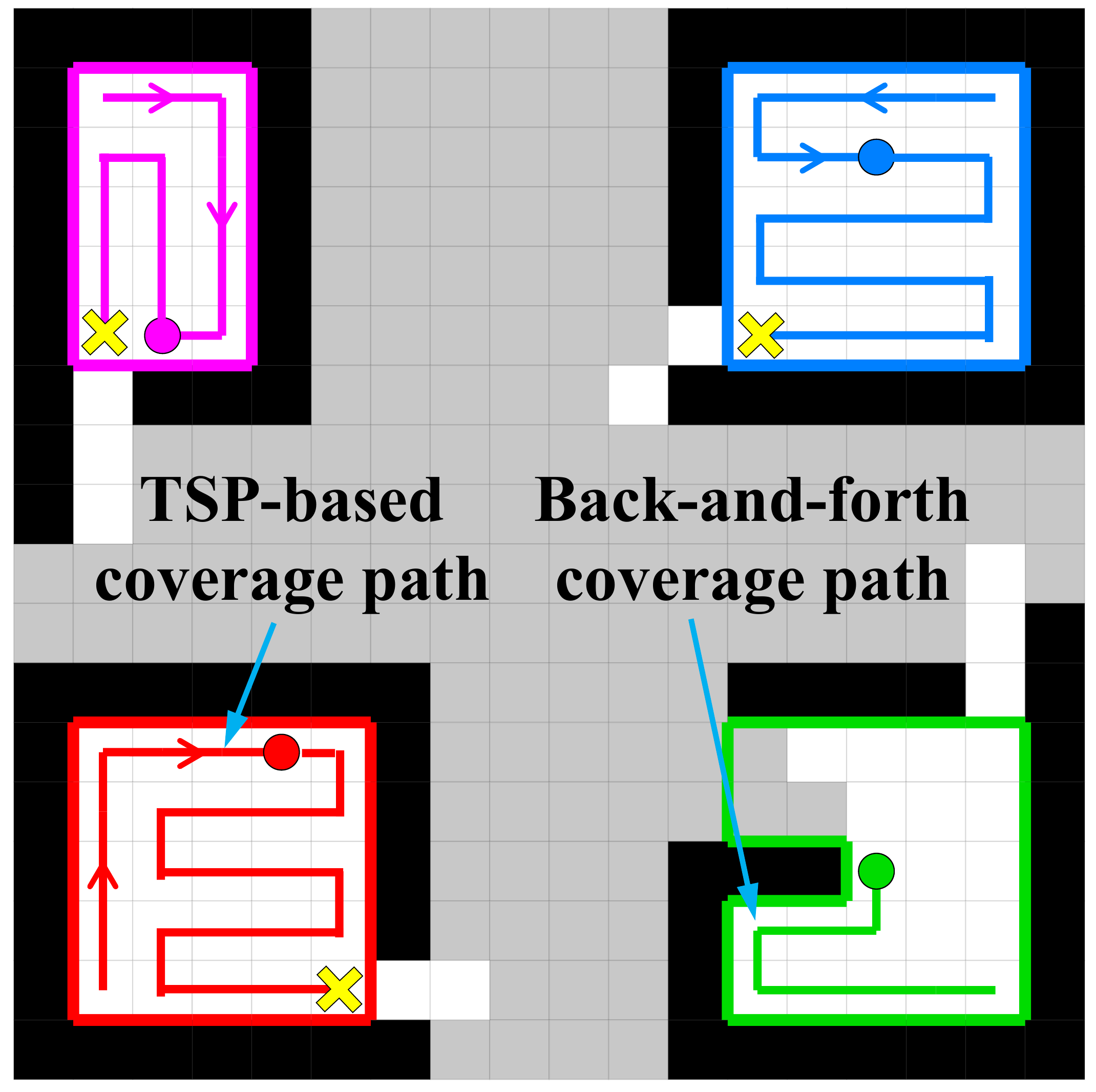}\label{fig:example_part4}}\vspace{4pt}\quad

    \centering
    \subfloat{
    \includegraphics[width=0.99\textwidth]{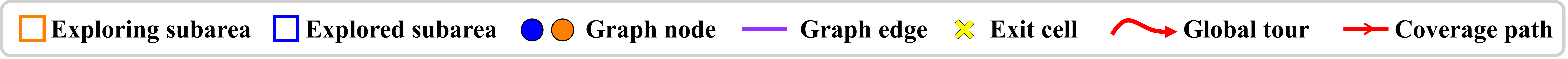}\label{fig:legend}}\\

    \caption{Illustration of the Multi-CAP algorithm. (a) The coverage coordination graph is initialized by partitioning the environment into a coarse uniform grid. Each coarse cell defines an initial subarea corresponding to a graph node, with edges representing adjacency between neighboring subareas. (b) The graph is incrementally updated using sensor feedback: blocked edges are removed, obstacle cells are excluded from subareas, and disconnected subareas are split into new ones with updated edges. (c) Global tours are computed over the graph to ensure all uncovered subareas are collectively visited. (d) Each robot independently adapts its coverage strategy: using a back-and-forth pattern in \textit{exploring} subareas and switching to a TSP-based shortest path in \textit{explored} subareas.}\label{fig:example} 
    \vspace{-1em}
 \end{figure*}

Let $\{r_m\}_{m=1}^{M}$ be $M \in \mathbb{N}^+$ robots, equipped with:
\begin{itemize}
\item A range detector (e.g., LiDAR) for mapping, 
\item A localization device (e.g., IMU),
\item A coverage tool or sensor (e.g., cleaning brush), and
\item A communication module.
\end{itemize} 

We assume a base station that collects sensor data from all robots and coordinates the coverage process. Let $\mathcal{X} \subset \mathbb{R}^2$ be a bounded, initially unknown environment. For mapping and coverage, we construct a uniform tiling over $\mathcal{X}$ as follows.

\begin{defn}[Uniform Tiling]\label{define:Tiling}
A set $\mathcal{T} = \{\tau_{i} \subset \mathbb{R}^2 : i=1,\ldots,I\}$ is called a uniform tiling of $\mathcal{X}$, where $I$ is the total number of cells, if:
\begin{itemize}
\item Each $\tau_{i}$ is a square cell of identical size,
\item $\tau_{i} \cap \tau_{i'} = \emptyset$ for all $i \neq i'$, 
\item $\mathcal{X} \subseteq \bigcup_{i=1}^{I} \tau_{i}$.
\end{itemize}
\end{defn}

Let $\mathcal{T}_{O} \subset \mathcal{T}$ denote the set of obstacle cells. The set of free cells to be covered is then given by $\mathcal{T}_{F} = \mathcal{T} \setminus \mathcal{T}_{O}$. The objective of our algorithm is to achieve complete coverage of $\mathcal{X}$, as formalized below.

\begin{defn}[Complete Coverage]\label{define:completeness} 
Let $\tau_m(k) \in \mathcal{T}_{F}$ denote the cell visited by robot $r_m$ at time $k$. Coverage of $\mathcal{X}$ is said to be \emph{complete} if there exists $K \in \mathbb{Z}^+$ such that the sequences $\{\tau_m(k) : k=1,\ldots,K\}_{m=1}^{M}$, taken over all robots, jointly cover all free cells, i.e., $\mathcal{T}_{F} \subseteq \bigcup_{m=1}^M \bigcup_{k=1}^K \tau_m(k)$.
\end{defn}

\section{Multi-CAP Algorithm}\label{sec:algorithm}
This section presents the Multi-CAP algorithm, which adaptively generates coverage paths for multiple robots using real-time sensor feedback in three stages. First, the base station aggregates sensor data to update the adjacency graph, capturing the environment’s connectivity and topology (Sec.~\ref{graphConstruct}). Next, the global planner assigns subarea tours by solving a modified Vehicle Routing Problem (Sec.~\ref{globalTourPlanning}). Finally, guided by the global tour, the local planner computes adaptive paths to cover each subarea (Sec.~\ref{localCover}). This process repeats until the environment is fully covered.

\subsection{Adjacency Graph Construction} \label{graphConstruct}
The adjacency graph provides a high-level topological abstraction of the environment, enabling structured monitoring of coverage progress and supporting both global coordination and local decision-making. It is an graph $\mathcal{G} = (\mathcal{N}, \mathcal{E})$, where:

\begin{itemize}
\item $\mathcal{N} = \{n_i: i = 0,\ldots,|\mathcal{N}|-1\}$ is the node set. Each node $n_i$ stores its associated subarea $\mathcal{A}(n_i)$, center position $p(n_i)$, and symbolic state $\mathcal{Q}(n_i)$, which encodes its observation and coverage status (see Defn.~\ref{stateencode}).

\item $\mathcal{E} = \{e_\alpha: \alpha = 0,\ldots,|\mathcal{E}|-1\}$ is the edge set. Each edge $e_{\alpha} \equiv e_{ii'}$ connects two adjacent nodes $n_i,n_{i'} \in \mathcal{N}$ and represents the collision-free path between their centers. 

\end{itemize}

\begin{defn}[State Encoding]
\label{stateencode}
For the purpose of tracking coverage progress, let $\mathcal{Q}: \mathcal{N} \rightarrow \{Cl,Op,\overline{Op}\}$ be a function that assigns a state $\mathcal{Q}(n_i)$ to each node $n_i \in \mathcal{N}$. The state $Cl$ indicates that the node has been covered by a robot. In contrast, $Op$ and $\overline{Op}$ denote uncovered nodes whose associated subareas are \textit{explored} and \textit{exploring}, respectively. 
\end{defn}

\begin{rem}
    A subarea is said to be \textit{exploring} if it contains unknown cell. Otherwise, it is classified as \textit{explored}.
\end{rem}

The graph is initialized by partitioning the environment into a coarse uniform grid, where each cell defines an initial subarea and is assigned to a graph node. Edges are added between neighboring cells to represent adjacency.

During online operation, the base station collects sensor data from robots and updates the occupancy probability of each cell $\tau_i \in \mathcal{T}$~\cite{thrun2005}. This probabilistic information is encoded in a symbolic map $\Phi: \mathcal{T} \rightarrow \{U,O,\widetilde{F},\hat{F}\}$. Cells that have not yet been observed are labeled $U$, while those with occupancy probability greater than $0.2$ are labeled $O$. The remaining cells are navigable; among these, those already-covered are labeled $\widetilde{F}$, and uncovered cells are labeled $\hat{F}$.

Given the updated symbolic map, the adjacency graph is incrementally refined to preserve two key properties: 1) each subarea remains a connected component of non-obstacle cells within its original region, and 2) each edge represents a valid, collision-free path between adjacent subareas. The complete update procedure is detailed below.

For each node $n_i$, the update procedure depends on whether it is affected by environmental changes. If unaffected, edge validity is checked, and any blocked edge is updated by recomputing a collision-free path using A* algorithm~\cite{hart1968formal}. If the node is affected, its subarea $\mathcal{A}(n_i)$ is re-evaluated using a flood-fill algorithm. Starting from an unexamined non-obstacle cell $\tau_i \in \mathcal{A}(n_i) \setminus \mathcal{T}O$, a recursive search expands in the four cardinal directions, labeling all connected non-obstacle cells with index $\ell \in \mathbb{N}^+$. This continues until expansion halts at boundaries or obstacles. Each labeled region defines a connected component $\mathcal{A}_\ell \subseteq \mathcal{A}(n_i)$, and the process repeats until all valid cells are examined. The result is a set of components $\{\mathcal{A}_\ell\}_{\ell=1}^{L}$, the node $n_i$ is retained, and its edges and state are updated. If $L > 1$, the subarea has fragmented; $n_i$ and its incident edges are removed, and new nodes are created for each connected component with associated edges. This procedure is applied to all nodes, ensuring that the graph evolves consistently with the environment and remains valid for hierarchical planning.

Figs.~\ref{fig:example_part1} and~\ref{fig:example_part2} illustrate this process. In Fig.~\ref{fig:example_part1}, the graph is initialized using a coarse grid of nine cells, each corresponding to a graph node. Edges connect neighboring subareas. In Fig.~\ref{fig:example_part2}, the symbolic map reflects newly detected obstacles that affect all existing nodes. Obstacle cells are removed from the subareas of nodes $\{n_0, \ldots, n_7\}$, and their states are updated. The edge between $n_0$ and $n_5$ is blocked and removed. The subarea of node $n_8$ is split into disconnected regions; thus, $n_8$ and its edges are removed, and new nodes are created for each resulting component.

\subsection{Global Tour Planning}
\label{globalTourPlanning}

Given the updated adjacency graph $\mathcal{G} = (\mathcal{N}, \mathcal{E})$, the objective is to compute tours for a team of robots such that all uncovered nodes are collectively visited. The goals are twofold: 1) minimize the total traversal cost and 2) ensure that robot tours are disjoint. Unlike the standard Vehicle Routing Problem (VRP), which assumes a single depot and closed-loop tours, our formulation allows multiple starting positions and open routes that do not require returning to the origin. Accordingly, the problem is cast as an Open Multi-Depot VRP (Open MDVRP).

Let $\{\bar{r}_m\}_{m=1}^{M'}$ denote the team of $M'$ robots available for new assignments, with current positions $p_{\bar{r}_m}$. The set of candidate nodes is $\mathcal{N}_{\text{cand}} := \{ n_i \in \mathcal{N} \mid \mathcal{Q}(n_i) \neq Cl \}$, representing all uncovered nodes. From this, we construct an augmented node set $\hat{\mathcal{N}}$, where the first $M'$ nodes correspond to robot positions $\{p_{\bar{r}_m}\}_{m=1}^{M'}$ and the remainder correspond to $\mathcal{N}_{\text{cand}}$. For each pair $\hat{n}_i, \hat{n}_j \in \hat{\mathcal{N}}$, the cost $w_{ij}$ is the length of the shortest collision-free path between them. To reformulate the problem as a standard VRP, we introduce a virtual depot $\hat{n}_{\hat{\mathcal{N}}}$. The cost between the virtual depot and each robot’s current position is set to zero, while the costs between the virtual depot and all other nodes are set to infinity. The resulting cost matrix is $\mathcal{W}=\left[w_{ij}\right]_{(\hat{\mathcal{N}}+1) \times (\hat{\mathcal{N}}+1)}$, with $w_{ii} = 0, \forall i\in \{0,\ldots,|\hat{\mathcal{N}}|\}$. 

Let $\Lambda$ be the set of feasible multi-robot solutions over $\mathcal{W}$. Each solution $\Pi\in\Lambda$ consists of $M'$ tours $\Pi = \{\pi_m\}_{m=1}^{M'}$, where each tour is an ordered sequence of nodes:
\begin{equation}\label{eq:ogm_recursive}
\pi_m=\{n(k) \in \hat{\mathcal{N}}: k=0,\ldots,K_m\}
\end{equation} 
where $n(k)$ is the node visited at step $k$ on the route $\pi_m$ by robot $r_m$, and $n(k)\neq n(k')$, $\forall  k,k'\in \{0,\ldots,K_m-1\}, k\neq k'$. Also, the first $n(0)$ and last element $n(K_m)$ represent the virtual depot $\hat{n}_{\hat{\mathcal{N}}}$. The total cost of solution $\Pi$ is given by:
\begin{equation}
\mathcal{J}(\Pi)= \sum_{m=1}^{M'}\sum_{k=0}^{K_m-1}w_{n(k)n(k+1)}
\end{equation}
The optimal solution $\Pi^{*}$ is then defined as:
\begin{equation}
\Pi^{*} = \mathop{\argmin}_{\Pi \in \Lambda}\mathcal{J}(\Pi)
\end{equation}  

Since the Open MDVRP is NP-hard, we adopt a two-stage heuristic. First, a nearest-neighbor algorithm incrementally assigns candidate nodes to robots based on spatial proximity. Second, each tour is refined using the 2-opt heuristic~\cite{aarts2003}. Each robot’s final tour $\pi_m$ is obtained by removing the virtual depot and its current position node, resulting in an open route visiting its assigned targets. The first element in the route is selected as the robot’s next target $n_m^{\text{target}}$ for local coverage. Fig.~\ref{fig:example_part3} illustrates an example of the resulting global tours.

\subsection{Local Coverage Path Planning}
\label{localCover}
The local planner operates in a distributed manner, allowing each robot to independently compute a coverage path within target subarea $\mathcal{A}(n_m^{\text{target}})$. The planning strategy depends on the state of target node $\mathcal{Q}(n_m^{\text{target}})$ and adapts accordingly to optimize coverage performance. Fig.~\ref{fig:example_part4}  illustrates an example of the resulting local coverage paths.

\begin{itemize}
    \item \textbf{Case 1:} If $\mathcal{Q}(n_m^{\text{target}}) = \overline{Op}$, robot $r_m$ follows a back-and-forth coverage pattern. At each step, it selects an uncovered cell within a $3\times3$ local grid, prioritizing directions in the global frame in the order: Left, Up, Down, Right. This rule produces a systematic back-and-forth path with a left-to-right sweep. If no uncovered cell is available locally, the robot moves to the nearest uncovered cell within the subarea $\mathcal{A}(n_m^{\text{target}})$. Since each subarea forms a connected component, the nearest uncovered cell is always within a bounded distance. Each visited cell is marked as covered. This continues until all free cells in $\mathcal{A}(n_m^{\text{target}})$ are covered. Once complete, the node $n_m^{\text{target}}$ is marked as \textit{Cl}.
    
    \item \textbf{Case 2:} If $\mathcal{Q}(n_m^{\text{target}}) = Op$, robot $r_m$ computes a shortest open-loop coverage path by solving a Traveling Salesman Problem (TSP) over all uncovered cells in $\mathcal{A}(n_m^{\text{target}})$, starting from its current position. Let $n_m^{\text{next}} \in \pi_m$ denote the next node in the global tour. If it exists, the exit cell is selected to enable a smooth transition to $\mathcal{A}(n_m^{\text{next}})$ by identifying a free cell along the transition path (from the global cost matrix $\mathcal{W}$) that borders the adjacent subarea. If no such node exists, the exit cell remains unspecified. To enforce start and end constraints, a dummy node is introduced with zero cost to both the start and exit cells, and infinite cost to all others. The TSP is solved using the 2-opt heuristic~\cite{aarts2003}. The resulting path is then executed by robot $r_m$, completing coverage of $\mathcal{A}(n_m^{\text{target}})$ and updating the node state to \textit{Cl}.
\end{itemize}

\begin{figure}[b]
    \centering
    \includegraphics[width=0.98\columnwidth]{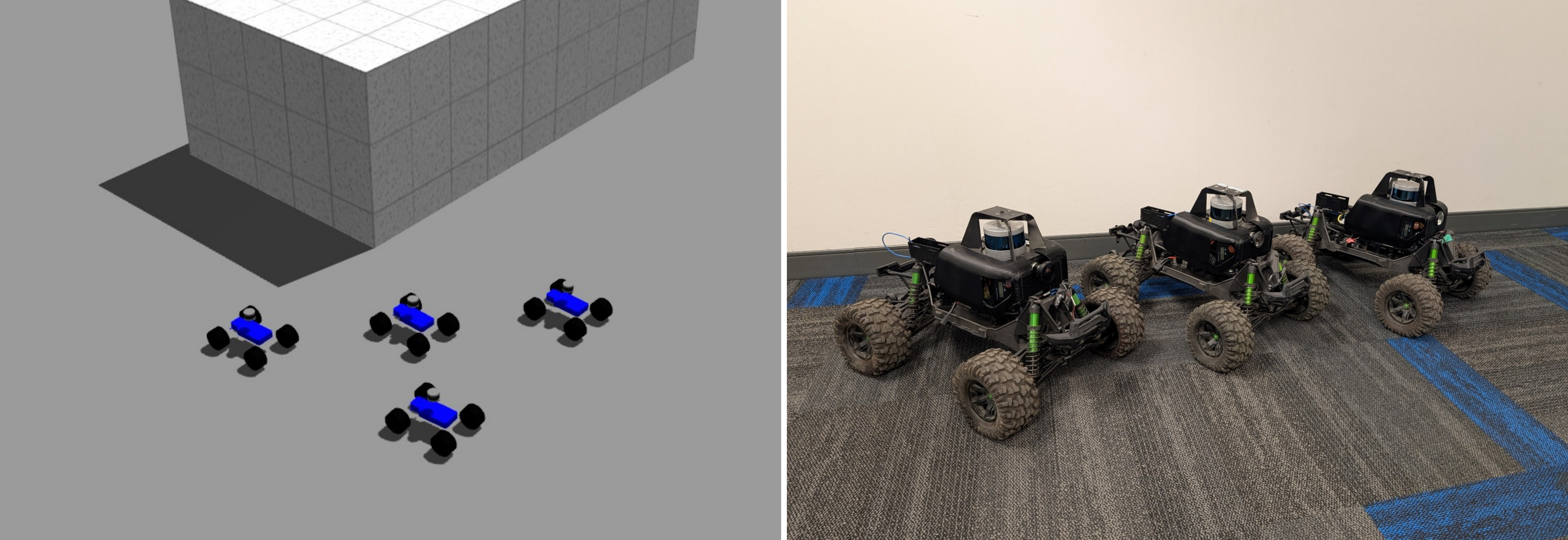}
    \caption{Robot platforms for simulation and real-world experiments.}
  \label{fig:robot}
  \vspace{-1.0em}
\end{figure}

\begin{figure*}[t]
    \centering
    \subfloat[Scene 1: Gallery.]{
    \includegraphics[width=0.99\textwidth]{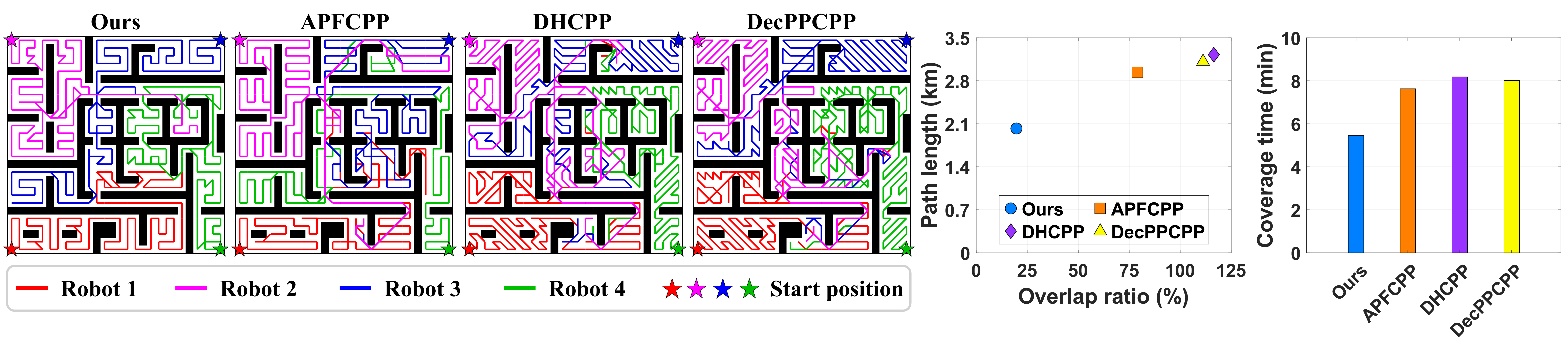}\vspace{-1.2em}\label{fig:simulation_scenario1}}\quad
    \centering
    \subfloat[Scene 2: Warehouse.]{
    \includegraphics[width=0.99\textwidth]{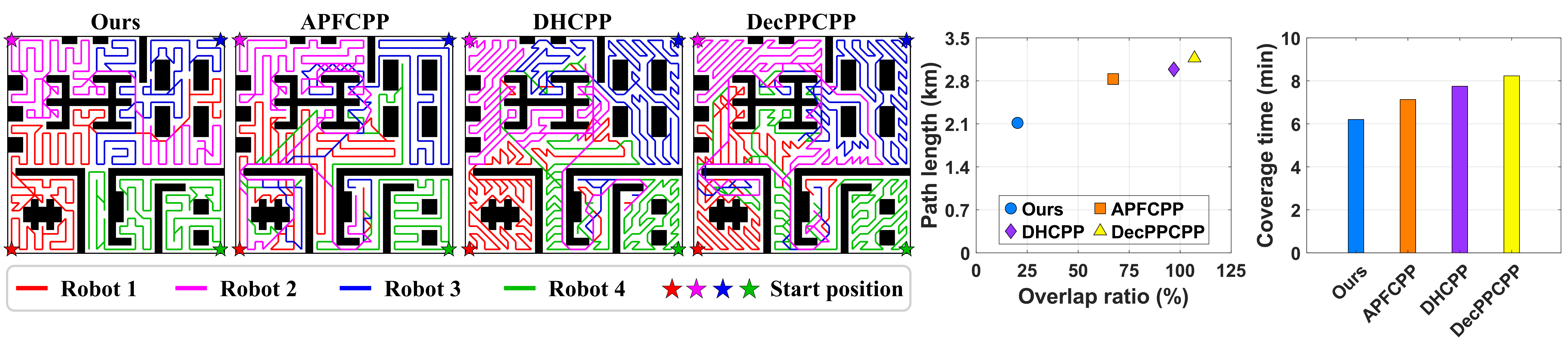}\vspace{-1.2em}\label{fig:simulation_scenario2}}\quad
    \centering
    \subfloat[Scene 3: Office.]{
    \includegraphics[width=0.99\textwidth]{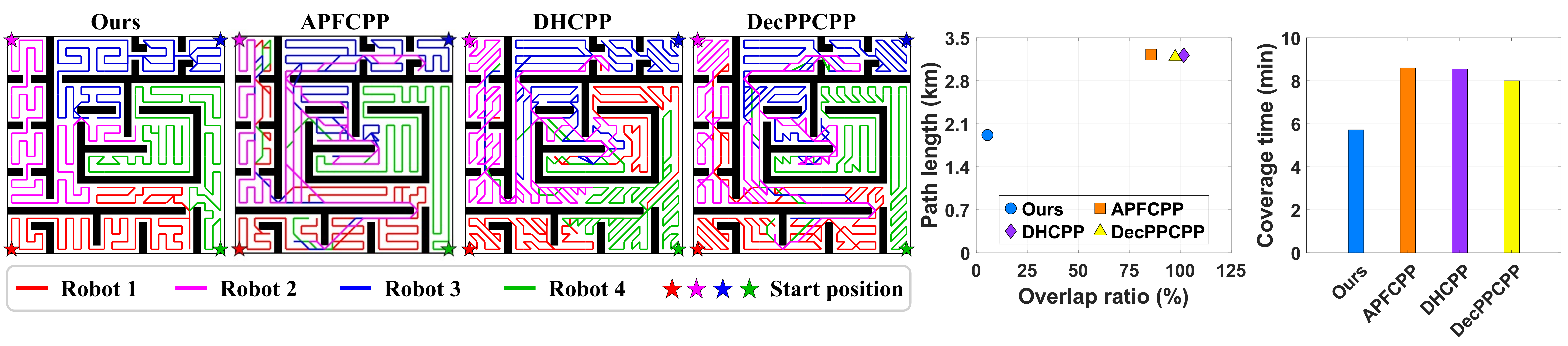}\vspace{-1.2em}\label{fig:simulation_scenario3}}\quad
    \centering
    \subfloat[Scene 4: Maze.]{
    \includegraphics[width=0.99\textwidth]{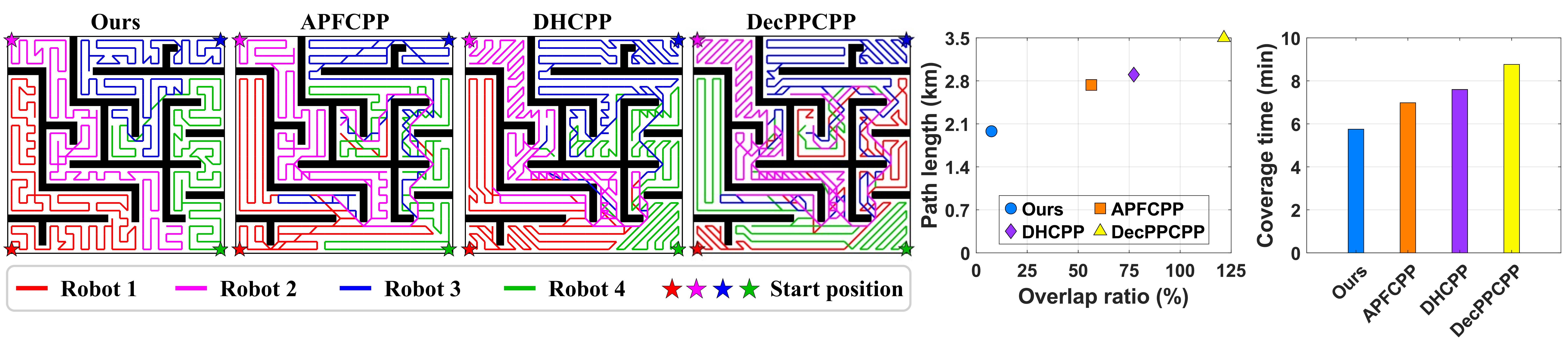}\vspace{-1.2em}\label{fig:simulation_scenario4}}\\
    \caption{Simulation experiments for Multi-robot CPP in four scenes.}
  \label{fig:simulation_result}
  \vspace{-1.0em}
\end{figure*}

\subsection{Time Complexity Analysis}
We analyze the time complexity of three core steps: adjacency graph construction (Sec.~\ref{graphConstruct}), global tour planning (Sec.~\ref{globalTourPlanning}), and local coverage path planning (Sec.~\ref{localCover}).  

Each affected subarea is re-evaluated using a flood-fill algorithm, which visits each cell at most once, giving $O(|\mathcal{T}|)$ complexity. The graph is then updated, with the main cost arising from A$^*$ searches used to compute edge paths. In the worst case, all edges are invalidated, requiring A$^*$ to run for all candidates in $\mathcal{E}$, where $|\mathcal{E}|\leq (|\mathcal{N}|^2 - |\mathcal{N}|)/2$. Since each A$^*$ search has complexity $O(|\mathcal{T}|\log|\mathcal{T}|)$ and $|\mathcal{N}|^2 \ll |\mathcal{T}|\log|\mathcal{T}|$, the overall complexity of graph construction is $O(|\mathcal{T}|\log|\mathcal{T}|)$. The global planner solves a Vehicle Routing Problem over $|\hat{\mathcal{N}}| + 1$ nodes using the heuristic approach, with complexity $O((|\hat{\mathcal{N}}| + 1)^2)$~\cite{shen2022ct}. The local planner operates independently for each robot, generating either a back-and-forth path with complexity $O(1)$ or a TSP-based shortest path with complexity $O(|\mathcal{A}(n_m^{\text{target}})|^2)$~\cite{shen2022ct}. Thus, the total time complexity is $O(|\mathcal{T}|\log|\mathcal{T}| + (|\hat{\mathcal{N}}| + 1)^2 + |\mathcal{A}(n_m^{\text{target}})|^2)$. Since $|\hat{\mathcal{N}}|^2 \ll |\mathcal{T}|\log|\mathcal{T}|$ and $|\mathcal{A}(n_m^{\text{target}})| \ll |\mathcal{T}|$, the overall complexity is $O(|\mathcal{T}|\log|\mathcal{T}|)$.

\section{Simulation Experiments}
\label{sec:simulation}

This section presents a comprehensive evaluation of the proposed algorithm through high-fidelity simulations conducted in the Gazebo environment. We assess both the overall performance and the contribution of individual components via comparative experiments against baseline methods and detailed ablation studies. All simulations are executed on a computer with a $2.60\,\mathrm{GHz}$ processor and $32\,\mathrm{GB}$ of RAM.

\begin{figure*}[t]
    \centering  
    \includegraphics[width=0.99\textwidth]{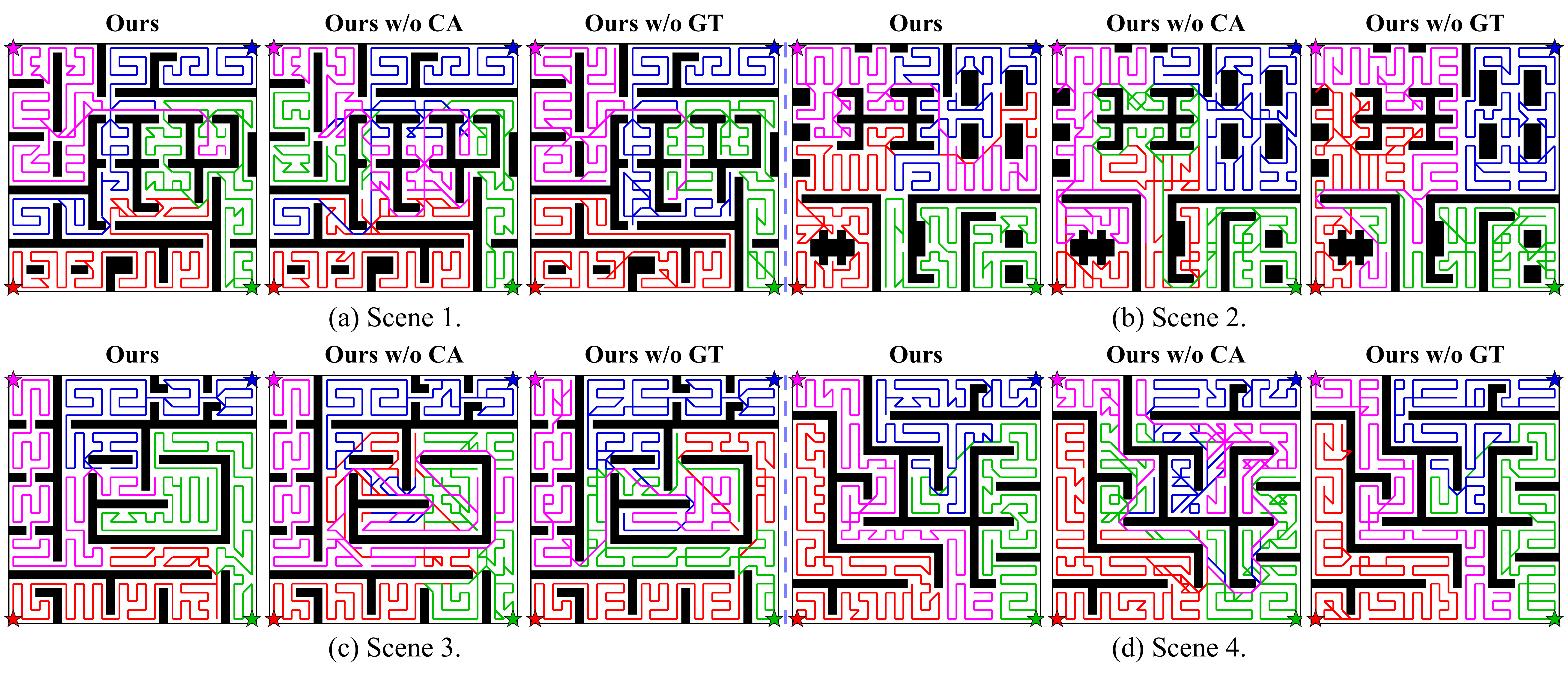}\vspace{-5pt}
    \caption{Coverage paths generated by the full algorithm and two baseline variants across four simulated scenes.}
  \label{fig:ablationstudy_path}
  \vspace{-1.0em}
\end{figure*}

\begin{figure*}[t]
    \centering
    \includegraphics[width=0.95\textwidth]{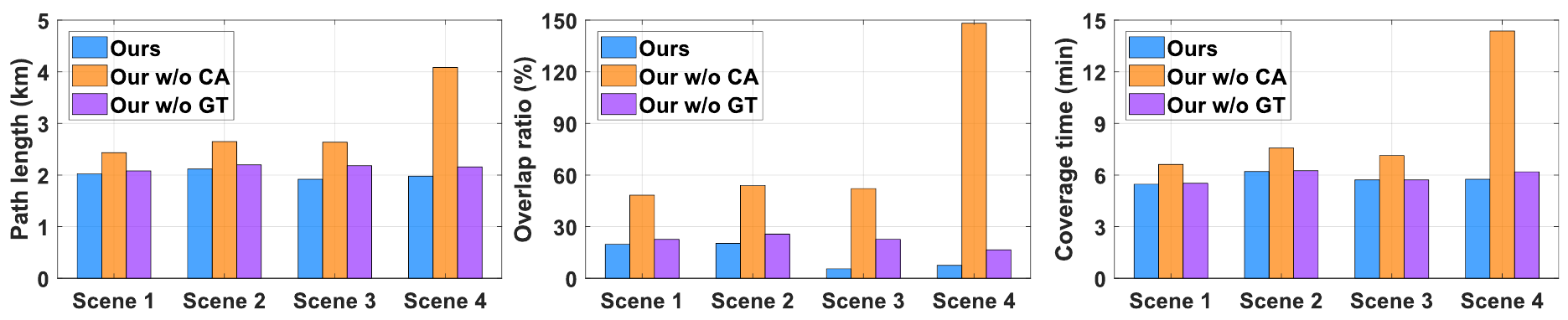}
    \caption{Quantitative results of the ablation studies on connectivity-aware decomposition and global tour planning.}
  \label{fig:ablationstudy_metric}
  \vspace{-1.0em}
\end{figure*}

\begin{figure*}[t]
    \centering
    \subfloat[Coverage paths generated by different algorithms in scene 1.]{
    \includegraphics[width=0.95\textwidth]{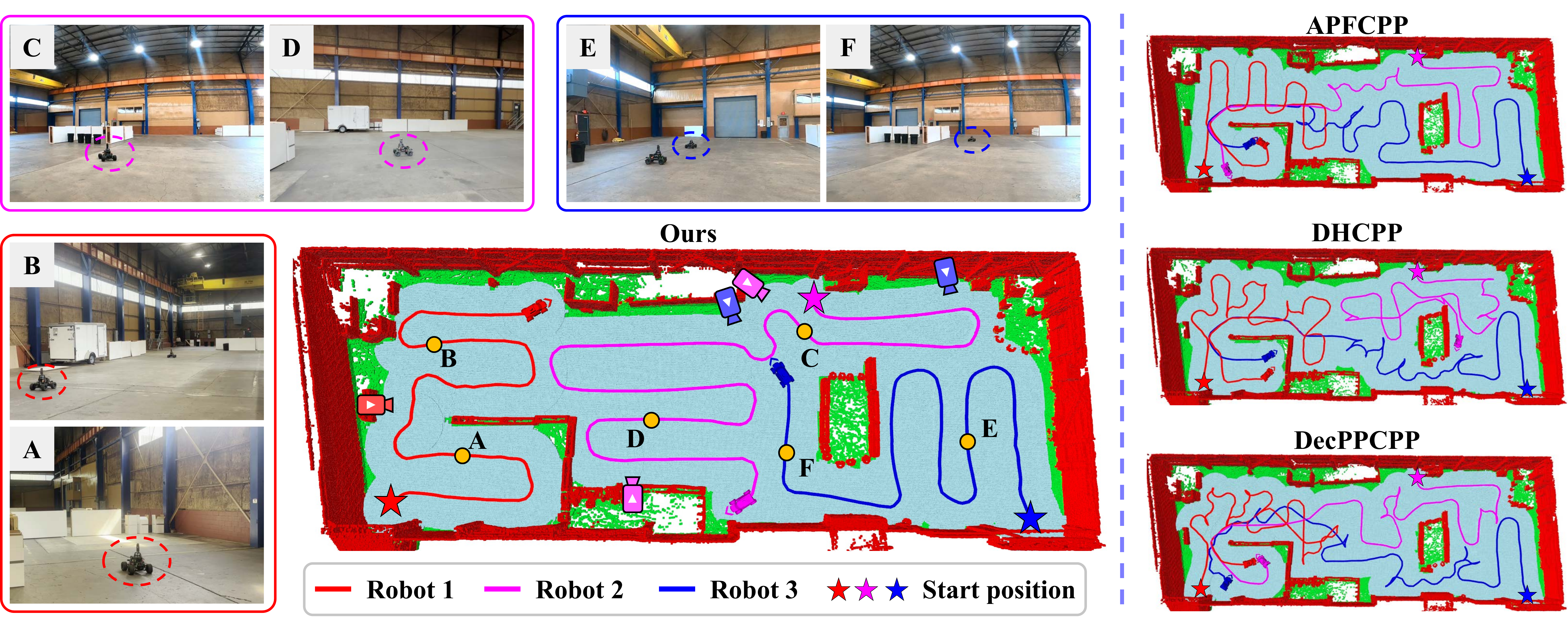}\label{fig:experiment_multiRobot_scenario1}}\vspace{0.2em}\quad
    \centering
    \subfloat[Coverage paths generated by different algorithms in scene 2.]{
    \includegraphics[width=0.95\textwidth]{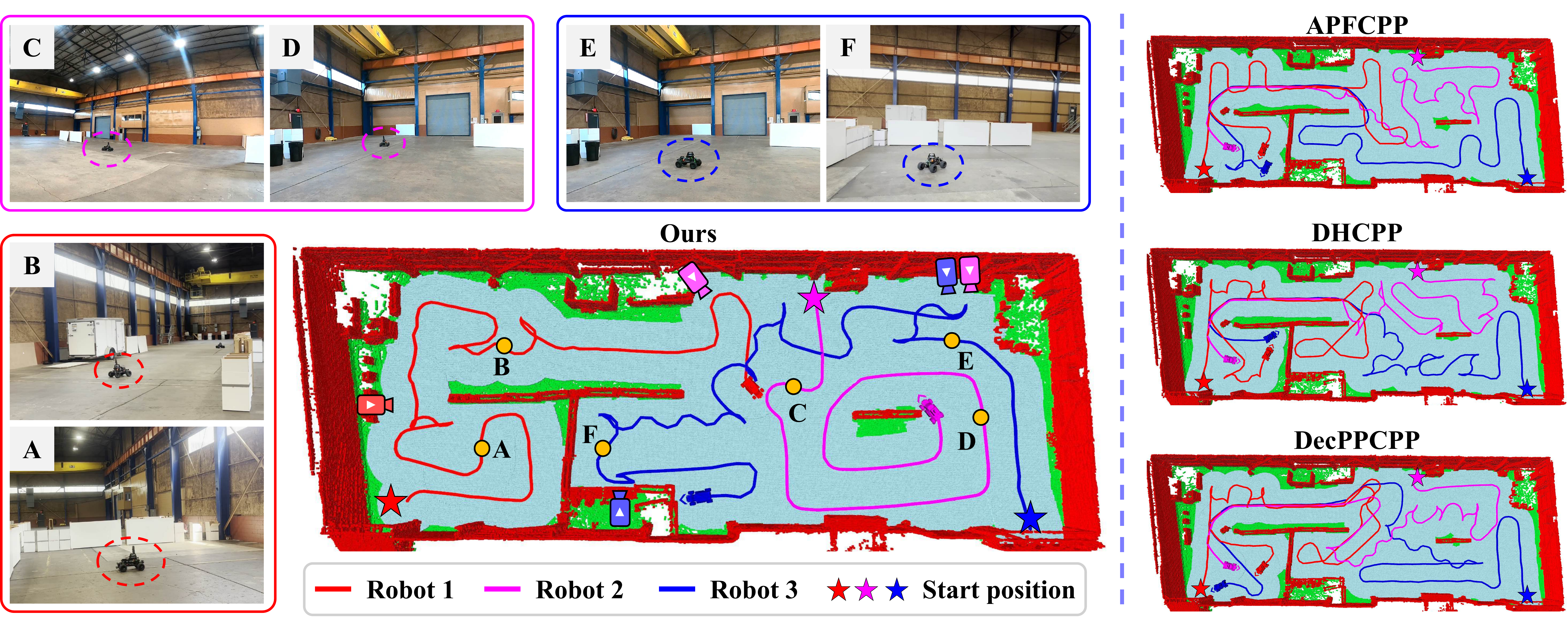}\label{fig:experiment_multiRobot_scenario2}}\vspace{0.2em}\quad
    \centering
    \subfloat[Quantitative results showing path length, overlap ratio, and coverage time.]{
    \includegraphics[width=0.99\textwidth]{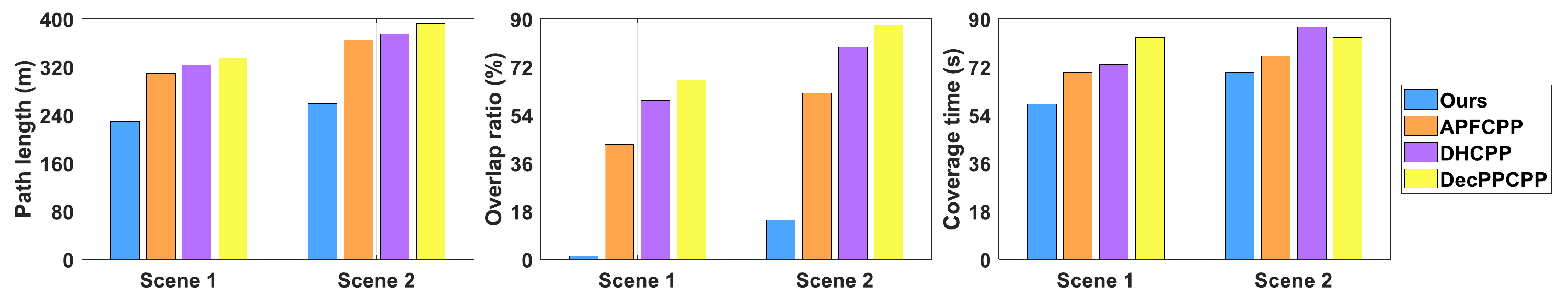}\label{fig:metric_experiment}}\\
    \caption{Real-world experiments for Multi-robot CPP in two scenes.}
  \label{fig:experiment_result}
  \vspace{-1.0em}
\end{figure*}

\subsection{Simulated Robot and Test Scenarios}
Fig.~\ref{fig:robot} shows the simulated wheeled robot platform, equipped with a Velodyne VLP-16 LiDAR providing a $360^\circ$ field of view and a maximum range of $12\,\mathrm{m}$. The robot adheres to realistic motion constraints, including a maximum speed of $2\,\mathrm{m/s}$ and a minimum turning radius of $1\,\mathrm{m}$. To evaluate robustness and generalizability, we test our algorithm across four simulated scenarios reflecting real-world challenges such as large-scale layouts, dense obstacles, and complex spatial topologies (see Fig.~\ref{fig:simulation_result}). The scenarios include a mall, warehouse, office, and maze, each spanning a $90\,\mathrm{m} \times 90\,\mathrm{m}$ area, discretized into a $30 \times 30$ tiling for mapping and coverage. Subareas are initialized with size $6\,\mathrm{m} \times 6\,\mathrm{m}$.

\subsection{Baseline Algorithms and Evaluation Metrics} 
We compare our algorithm against DecPPCPP~\cite{hassan2020dec}, APFCPP~\cite{wang2024apf}, and DHCPP~\cite{zhang2024herd}. To resolve local conflicts in all methods, a centralized multi-robot path planning framework~\cite{Yu2016} is employed. The performance metrics are:
\begin{itemize}
    \item \textbf{Path Length:} The total distance traveled by all robots to achieve complete coverage.
    \item \textbf{Overlap Ratio:} The proportion of repeatedly covered free cells relative to the total number of free cells.
    \item \textbf{Coverage Time:} The total operation time of the robot team, determined by the completion time of the last robot to finish its coverage task.
\end{itemize}

\subsection{Results and Discussion}

Fig.~\ref{fig:simulation_result} shows the coverage paths generated by our algorithm and the three baseline algorithms across four different scenarios, where all robots start at the corners of the area. During online operation, the robots incrementally discover the environment and adapt their coverage paths using each algorithm until the entire area is covered. As observed, our algorithm consistently produces more efficient coverage paths with significantly less overlap compared to the baselines. Quantitative comparisons in terms of path length, overlap ratio, and coverage time are also presented in Fig.~\ref{fig:simulation_result}. Across all metrics and scenarios, our method demonstrates clear performance advantages. Specifically, when compared to the second-best performing baseline in each scenario, our algorithm achieves reductions in coverage time by $28\%$, $13\%$, $29\%$, and $18\%$, respectively. These improvements can be attributed to the following key design features:

\begin{itemize}
    \item \textbf{Connectivity-Aware Space Decomposition:} The environment is partitioned into distinct subareas, ensuring that each subarea forms a connected component. This avoids long detours within a subarea during local coverage. Moreover, adjacency is defined only when the union of two subareas is path-connected, enabling a more coherent and efficient global tour.

    \item \textbf{Global Tour Planning:} Robots follow a globally optimized tour over uncovered subareas, minimizing unnecessary transitions. By assigning distinct subareas to different robots, local conflicts are reduced.
    
    \item \textbf{Adaptive Local Coverage Path Planning:} The local planner uses a back-and-forth pattern for \textit{exploring} subareas, while switching to a shortest coverage path for \textit{explored} subareas. This hybrid strategy effectively reduces overall path length and redundant travel.
\end{itemize}

Among the baselines, APFCPP performs relatively better due to its use of repulsive fields that promote spatial dispersion and reduce local conflicts. However, because it allows all robots to operate across the entire area without explicit spatial allocation, it still suffers from coverage inefficiencies.

\subsection{Ablation Studies}
\label{ablation}
We conduct ablation studies to evaluate the effectiveness of each algorithmic component.

\subsubsection{Connectivity-Aware Space Decomposition}
We design a baseline variant named \textbf{Ours w/o CA}, which replaces the connectivity-aware decomposition method with a uniform decomposition
method. Specifically, the environment is simply divided into equal-sized subareas without considering connectivity. Moreover, this baseline defines subarea adjacency based solely on spatial proximity. As shown in Fig.~\ref{fig:ablationstudy_path}, the baseline generates noticeably more overlapping paths due to its lack of consideration for environmental topology and connectivity. This leads to long detours during local coverage, as individual subareas can be split into disconnected portions by obstacles. Additionally, global tours become inefficient since subarea adjacency is defined purely based on spatial proximity. In contrast, our full algorithm captures the structural properties of the environment, resulting in more coherent and efficient coverage. Fig.~\ref{fig:ablationstudy_metric} presents the quantitative results, where the full algorithm achieves substantial improvements across all metrics, demonstrating the effectiveness of the connectivity-aware decomposition.

\subsubsection{Global Tour Planning}
We design a baseline variant named \textbf{Ours w/o GT}, which selects the nearest uncovered subarea as the next target. Fig.~\ref{fig:ablationstudy_path} compares the coverage paths produced by the full algorithm and this variant. The baseline results in more overlapping paths because it ignores the global traversal order of subareas, leading to unnecessarily long transition paths. In contrast, the full algorithm covers each subarea by following a globally optimized sequence. As shown in Fig.~\ref{fig:ablationstudy_metric}, the full algorithm achieves clear improvements in path length and overlap ratio, while the coverage time remains similar. This is because coverage time is primarily determined by the workload of the slowest robot, which is not significantly altered by global sequencing. Thus, global tour planning improves efficiency by reducing redundant travel rather than accelerating task completion.

\section{Real-World Experiments}
\label{sec:experiment}

Our algorithm is further evaluated through real-world experiments conducted in two scenes, shown in Fig.~\ref{fig:experiment_multiRobot_scenario1}. Each scene spans $60\,\mathrm{m} \times 24\,\mathrm{m}$ and is discretized into a $20 \times 8$ grid for mapping and coverage. Subareas are initialized with size $6\,\mathrm{m} \times 6\,\mathrm{m}$. Three wheeled robots are deployed, each equipped with an NVIDIA Jetson AGX Orin, a Velodyne VLP-16 LiDAR, an EPSON-G366 IMU, and a radio communication device (see Fig.~\ref{fig:robot}). The robots operate under realistic motion constraints, with a maximum speed of $2\,\mathrm{m/s}$ and a minimum turning radius of $1\,\mathrm{m}$. For localization and mapping, we employ the LiPO SLAM algorithm~\cite{LiPo}, and the full autonomy stack is adapted from~\cite{sriganesh2024systemdesign}. The coverage paths generated by each algorithm are presented in Fig.~\ref{fig:experiment_multiRobot_scenario1} and Fig.~\ref{fig:experiment_multiRobot_scenario2}, where our method achieves efficient coverage with minimal overlap. Quantitative results in Fig.~\ref{fig:metric_experiment} further show that our algorithm consistently outperforms the baselines across all metrics, confirming its efficiency and practicality in real-world multi-robot coverage tasks.

\section{Conclusions and Future Work} \label{sec:conclusions}
We presented the Multi-CAP algorithm for multi-robot coverage in unknown environments. The method decomposes the space into an adjacency graph, assigns subareas via a global Open Multi-Depot VRP, and computes locally adaptive coverage paths. This hierarchical strategy reduces redundant travel and inter-robot conflicts by integrating global coordination with local execution. Simulation and real-world experiments demonstrate consistent improvements over baselines, with ablation studies confirming the value of each component. Future work will explore communication-constrained settings, non-holonomic robots~\cite{shen2019online}, and coverage of dynamic environments with moving obstacles~\cite{Shen_SMART2023}.


\balance
\bibliographystyle{IEEEtran}
\bibliography{reference}

\end{document}